# BTPK-based interpretable method for NER tasks based on Talmudic Public Announcement Logic

Yulin Chen[a]    Beishui Liao[a,1]    Bruno Bentzen[a]    Bo Yuan[a]    Zelai Yao[a]

Haixiao Chi[a]    Dov Gabbay[b]

[a] *Zhejiang University, Zheda Road 38, Hangzhou, 310028, China*

[b] *King's College London, Strand London WC2R 2LS, United Kingdom London*

**Abstract:** As one of the basic tasks in natural language processing (NLP), named entity recognition (NER) is an important basic tool for downstream tasks of NLP, such as information extraction, syntactic analysis, machine translation and so on. The internal operation logic of current name entity recognition model is black-box to the user, so the user has no basis to determine which name entity makes more sense. Therefore, a user-friendly explainable recognition process would be very useful for many people. In this paper, we propose a novel interpretable method, BTPK (Binary Talmudic Public Announcement Logic model), to help users understand the internal recognition logic of the name entity recognition tasks based on Talmudic Public Announcement Logic. BTPK model can also capture the semantic information in the input sentences, that is, the context dependency of the sentence. We observed the public announcement of BTPK presents the inner decision logic of BRNNs, and the explanations obtained from a BTPK model show us how BRNNs essentially handle NER tasks.

---

[1] Corresponding author

# 1 Introduction

Named Entity Recognition (NER) is an information extraction task aimed at classifying words in unstructured text [3,6]. Due to their ability to establish dependencies in neighboring words, BRNNs have demonstrated excellent performance in many NER tasks [9]. Despite the advantages of such deep learning based methods, their inherent black box nature makes them unable to explain decision results [5,10]. In application areas where NER technology provides extensive underlying support such as health-care or autonomous driving, a transparent internal decision system is critical for the system reliability and user trust. Many interpretable works have been carried out for RNN [7,11,8], However, there are few research efforts on explainability of BRNN in NER tasks, although models with explainability are crucial [2].

Talmudic public announcement logic (TPK) is one of the modal logic that serves as a good formalism for representing decisions depending on the future [1]. Regarding the past and future information as context information, TPK is able to model the context relationship and represent the implicit logic in context through modal logic. Some work has been done in exporting logical insights derived from the Talmud to AI, including the description of active historical databases, the study of retroactive update,etc. In this paper, we use the Talmudic public announcement logic (TPK) model [1] as a tool to explain the process of NER and bring transparency to the RNN-based models, since the reversible and modifiable recognition process in NER is very much in line with the problem that TPK is trying to deal with. We propose a new binary TPK model (called BTPK) based on the original TPK model, which can deal with actions depending on future determinations by public announcements [1]. Through modifying the accessibility relation in a temporal tree structure, the public announcement at a future state will tell which path should be chosen. Thus, a logical explanation can be obtained for any trained BRNN based on BTPK. We summarize our main contributions as follows: (1) We propose a BTPK-based learning method based on the original TPK model and apply it to a BRNN model to obtain logic explanationa for a BRNN. (2) We carry out a case study on real dataset to show how BRNN handle NER tasks, as well as to explore the potential for further reasoning on a BTPK model.

## 2 Preliminary

The original TPK model is a deterministic Talmudic **K** frame based on a time-action tree structure. The time-action model is a tree structure with a set of states: $S = \{s_0, s_1, s_2, s_2', \dots\}$ ( $s_0$ is the root), and a set of actions $A = \{a_1, a_2, a_3', \dots\}$. The elements of A are actions moving the agent from any state to a new one. This corresponds to a successor function $R_1$ (denoted by →), and can be written in the form $s_0 R_1 s_1$. A time-action sequence has the form of $s_0 a_1 a_2 \dots a_n$.

In scenarios where the present course of action is indeterminate and the subsequent state is uncertain, the concept of a public announcement is proposed as a means of resolving the ambiguity. This entails providing clarification regarding previous undetermined path and identifying the accurate successor of the decision point. A deterministic TPK model [1] can be defined as a 6-tuple $(S, R_1, R, \rho, s_0, \pi)$ where $(S, R_1, s_0)$ is a tree with root $s_0$ and successor relation $R_1$, R is the transitive closure of $R_1$, $\rho$ is the public announcement function and $\pi$ is an assignment for each atom q, such that $s \models q$ iff $s \in \pi(q)$. D is the distance from the root, and if $s_3' \rho s_3$ then $D(s_3) = D(s_3') + 1$.

The semantics for Deterministic TPK Model is as follows:

(1) As for the relation $R_1$:

$t \models \Box A$ iff $\forall s: tR_1 s \rightarrow s \models A$;

$t \models YA$ iff $\forall s: sR_1 t \rightarrow s \models A$; $t \models YA$.

(2) As for the relation $\rho$:

$t \models \boxminus A$ iff $\forall s: t\rho s \rightarrow s \models A$;

$t \models \mathbb{Y} A$ iff $\forall s: s\rho t \rightarrow s \models A$.

(3) $D_n$ is a time constant: $t \models D_n$ iff the distance of t from $s_0$ is n.

## 3 Approach

The overall framework is illustrated in Figure 1. Initially, a Bidirectional Recurrent Neural Network (BRNN) learner is trained on a set of training data to obtain a well trained BRNN model. The public announcements of the sequences are computed through an analysis of the backward and forward hidden states. Section 3.1 provides the definition of public announcement within the binary TPK model (BTPK). In Section 3.3, the BTPK models of BRNNs are generated.

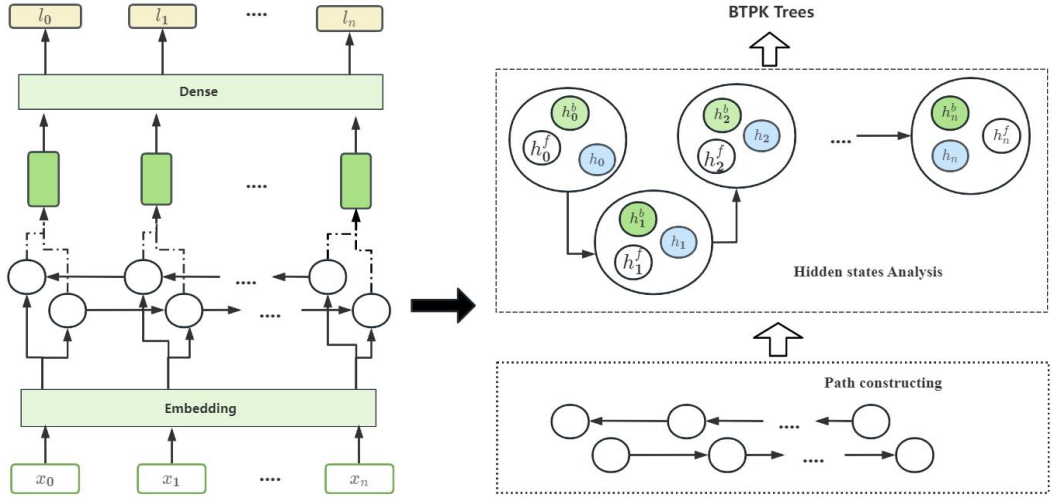

Figure 1 The illustration of the framework of BTPK-based learning. $h_i$ in the blue circles represents the a hidden state of in a BRNN, $h_i^f$ in the white circles represents the a hidden state in forward RNN and $h_i^b$ in the green circles represents a hidden state in backward RNN, where i is the element order. $l_i$ in blue box denotes the output of the BRNN.

## 3.1 Definitions

**Definition 3.1 (Task definition)** *We regard NER as a sequence labeling problem, whose input includes a set of sequences and labels. For any sequence W=($w_1, w_2, ... w_n$), the corresponding labels are Y=($y_1, y_2, ... y_n$), where $w_i$ denotes an entity in the sequence, and $y_i$ comes from BIO tagging schema for labeling elements from the sentence.*

We view the final output of one RNN as the final option, so there are at most two options for each entity in the BRNN. According to the original TPK model, we define a binary TPK logic (BTPK) model as follows.

**Definition 3.2 (BTPK)** *A binary TPK logic (BTPK) model is defined as a $T =< V, E >$ with public announcements $P$ and height |H|, where |V| is the order, $V = V_1 \cup V_2$, |E| is the size and E is*

*represented by the successor relation $R_1$. Height |H| denotes the depth of a tree.*

Based on above definitions, let each node of the same height be annotated by the possible label of a named entity, we can construct a tree which represents all options in BRNN to recognize a sequence.

## 3.2 Path construction

For any sequence $W=(w_1, w_2, ... w_n)$ and the corresponding labels $Y=(y_1, y_2, ... y_n)$, we can map the bi-directional hidden states to the path in a BTPK model $T' = <V, E>$, where the hidden states of BRNN constitute the vertices of BTPK. We present the mapping from a BRNN network to a BTPK model of height $|H| = n + 1$ as follows:

$$V_1 = \left\{h_1^f, h_2^f, ..., h_n^f\right\} \qquad V_2 = \left\{h_1^b, h_2^b, ..., h_n^b\right\}$$
$$H = \{w_0, w_1, w_2, ..., w_n\} \qquad n \leq |E| < |V| * (|V| - 1) \qquad (1)$$

where $h_i^f$ means the hidden state (feature vector) of the i-th element $w_i$ in forward RNN, and $h_i^b$ means the feature vector of the i-th element w_i in backward RNN, $w_0$ denotes the start. V is the vertices of graph, which is composed of $V_1$ and $V_2$, where $V_1$ and $V_2$ denote the vertices of forward RNN and backward RNN, respectively. H is denoted by the elements in the sequence. As mentioned above, for all x, y ∈ V, xy ∈ E iff $xR_1y$ or $yR_1x$, written as x < y or y < x. Unlike standard binary trees, the size |E| is greater than or equal to n because there may be loops in a BTPK model, since there exist public announcements in the tree.

It is important to note that the primary methodology for constructing the path of BTPK models involves the identification of branches, which entails recognizing the points in the model where different decisions may be made and subsequently indicating the potential outcomes of these decisions at the corresponding nodes. This study aims to identify the branches of trees in the BTPK model through the distinct masking of forward and backward hidden states.

## 3.3 Public announcement extraction

The present study involves the construction of the branch and path of BTPK models through the utilization of trained BRNNs. Nevertheless, the process of path construction solely achieves the representation of knowledge pertaining to concealed states of trained Bidirectional Recurrent Neural Networks (BRNNs), while the decision-making logic that is implicitly involved remains undisclosed. Consequently, we conduct a further examination of the correlation between these concealed states and utilize the technique of public announcement to represent said correlation.

The challenge of identifying public announcements includes identifying the key factor that determine the predicted label of a given entity. The public announcements in BTPK models are derived from the forward hidden states, backward hidden states, and BRNN hidden states that combine the forward and backward ones, as illustrated in Figure 1.

Firstly we split the input sequences into a series of grams such that:

$$G_i = \{w_j, \ldots w_k\} \qquad (0 \leq j \leq k \leq n) \qquad (2)$$

where $G_i$ denotes the i-th gram, and $n$ denotes the length of a input sequence. $w_j$ and $w_k$ denote the j-th words and k-th words in a input sequence, repectively. Then, we select grams one by one for intervention, where the selected gram will be mapped to the corresponding prototypes in the feature space, as shown in Figure 2. To be precise, the hidden states of the chosen n-grams are set to zero through physical intervention. The detection of public announcements is achieved through an analytical process that involves the examination of both the original hidden states and the invented hidden states, drawing inspiration from the techniques employed by the Millers' methods[4].

## 4 Experiments and Discussion

In this section, we will firstly introduce our experimental setup. Secondly, we will show how to generate a BTPK model from a trained BRNN by a real instance. Thirdly, we try to reason about ambiguous entities through TPK semantics.

## 4.1 Experiment setup

This paper train BRNNs on a chinese public NER dataset **CBVM** which is available on GITHUB and includes 7 label categories. We extract 8791 available sequences from it, including 7814 train samples and 977 test samples. As for training parameters, we set $batch\_size = 32$, $learning\_rate = 0.0001$. Hidden states and embedding dimensions are fixed at 128.

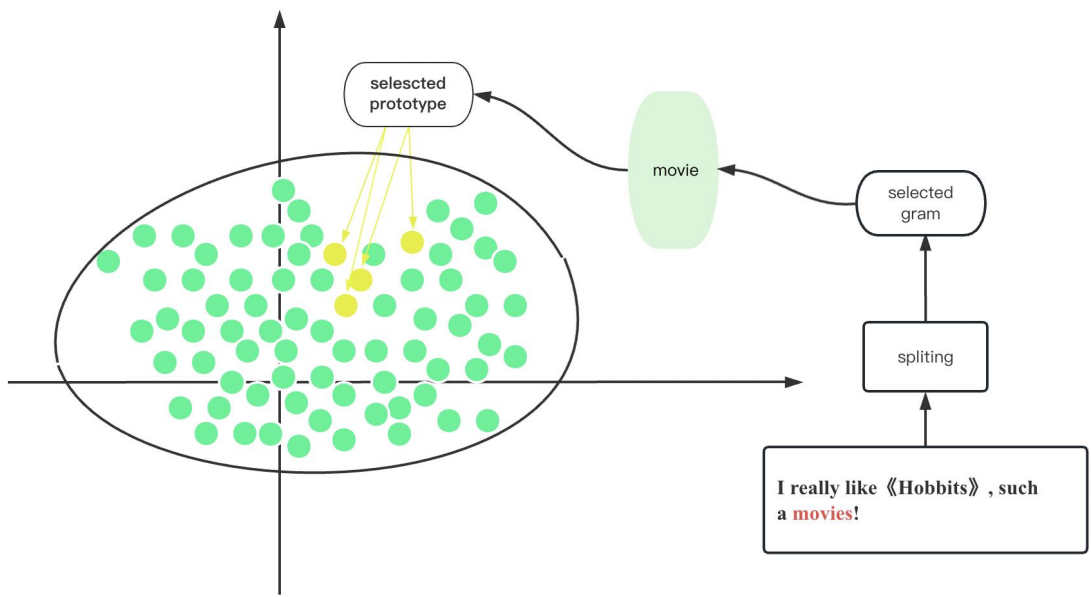

Figure 2. The illustration of the identification of the public announcement in *task103*.

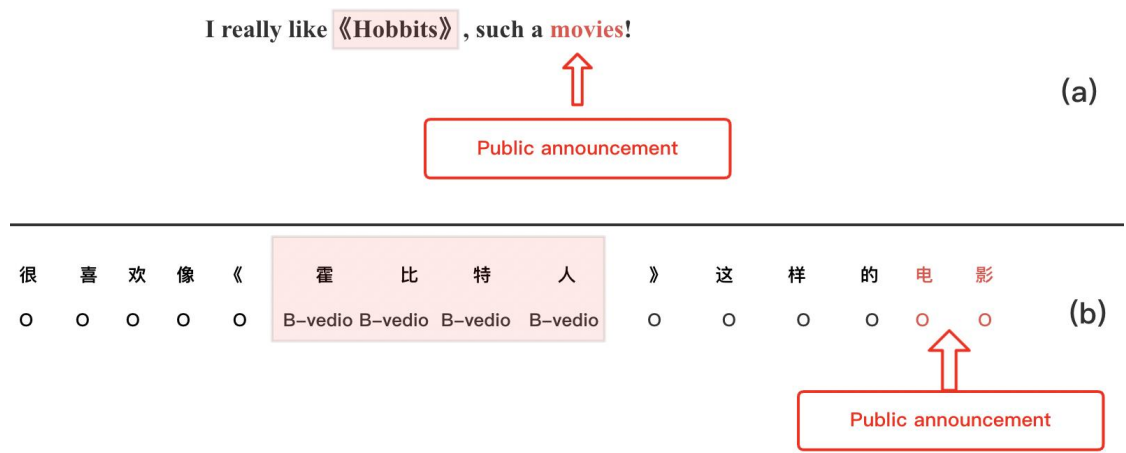

Figure 3. The illustration of the public announcement in *task103*. The original sequences are in Chinese, we also show it in English to help readers understand it.

## 4.2 Generating a BTPK model

**Example 4.1** *Consider the sentence task103 ={I really like "Hobbits", and movies like this". }*

For an arbitary input sequence, we can generate global explanation for the decision process of a trained BRNN by BTPK model. The hidden state analysis for **Example 4.1** is demonstrated in Figure 2. Firstly we split the input sequences into a series of grams and each gram is composed of one or more words. Then, the selected gram will be mapped to the corresponding prototypes in feature space, where the selected gram (words) are "movie" in Figure 2. According to our experimental results, we find that the predicted label of entities "Hobbits" will not be "video" name when we fix the hidden states of the selected n-grams to 0 by physical intervention. Therefore, it's possible to derive the conclusion that "movie" is the public announcement for the predicted label of entities "Hobbits", as shown in Figure 3. Based on above knowledge, a BTPK model is established in Figure 4.

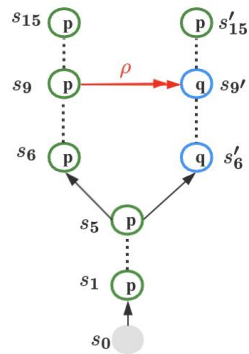

Figure 4: BTPK model.

## 4.3 Semantic ambiguity explanation

Intuitively, we consider the the human-readable explanations that consists of public announcements and a natural language template. To generate user-friendly explanations for those without background in name entity tagging, we consider "B_book" and "I_book" to be "book", "B_video" and "I_video" to be "video" and "B_music" and "I_music" to be "music". Generally, we have the following explanations for Example 4.1:

**Question**: Why is "Hobbits" recognized as a video name rather than other labels ( e.g., book name or music name)?

**Explanation**: Because the "movie " (public announcement) appears in the following words, it is more reasonable to be recognized as "video".

In this example, the explanation is obtained by a logic reasoning process of a BTPK model in Figure 4. Let axiom $q$ denotes that the entity is recognized as "vedio", axiom $p$ denotes that the entity is recognized as other labels. From semantics of TPK logic, "Hobbits" is correctly recognized iff $s_6' \models q$ and $s_7' \models q$ and $s_8' \models q$ and $s_9' \models q$, where $s_i'$ denote the states in height $|H| = i$. When the system gets words from $s_1$ and goes forward to $s_{15}$, the path can be represented as $s_1 \models \Box p$. But there is a public announcement $s_9 \rho s_9'$, so the system will go back to $s_6'$ and then go forward to the end state $s_{15}'$, generating a new path on the right branch of the tree. The new path can be denoted by ($s_6' \models q$ and $s_7' \models q$ and $s_8' \models q$ and $s_9' \models q$ and $s_5 \models Yp$ and $s_{10}' \models \Box p$ ), which is the ground truth of this sequence. Thus, the recognition process of the entity can be presented in a logical way by BTPK model, and the public announcement illustrates how to go back to a more reasonable state.

# 5 Conclusions and future work

We proposed a new BTPK-based interpretable method for NER tasks, which can effectively and logically capture the semantics in the context and give explanations in form of trees to show the internal mechanism of BRNN models. We implement the BTPK-based interpretable method to a trained BRNN model to obtain logic explanations. In addition, we also demonstrate how to reason on a BTPK model to understand the inner decision making path of a trained BRNN. For future work, we plan to combine BTPK-based interpretable method (as in this work) with transfer learning for cross-lingual NER tasks.